\documentclass[a4paper]{article}
\usepackage{INTERSPEECH2016}

\usepackage{amsmath,graphicx}
\usepackage{dblfloatfix} 
\usepackage{multirow}
\usepackage{tabularx}
\usepackage{array,booktabs}
\usepackage{threeparttable}
\usepackage{subfig}
\usepackage{url}
\usepackage{tipa}
\usepackage[svgnames]{xcolor}
\usepackage{mathrsfs} 
\usepackage{mathtools}
\usepackage{tensor}
\usepackage{listings}
\usepackage{color}
\usepackage{hyperref}
\ninept

\title{SPEECH-COCO: 600k Visually Grounded Spoken Captions Aligned to MSCOCO Data Set}
\makeatletter
\def\name#1{\gdef\@name{#1\\}}
\makeatother \name{{\em William N. Havard$^1$, Laurent Besacier$^1$, Olivier Rosec$^2$}}

\address{$^1$Univ. Grenoble Alpes, CNRS, Grenoble INP, LIG, F-38000 Grenoble, France \\
  $^2$Voxygen, France \\
  {\small \tt william.havard@gmail.com, laurent.besacier@univ-grenoble-alpes.fr, olivier.rosec@voxygen.fr} 
}

\begin{document}
%
\maketitle

\begin{abstract}

This paper presents an augmentation of MSCOCO dataset where speech is added to image and text. Speech captions are generated using text-to-speech (TTS) synthesis resulting in 616,767 spoken captions (more than 600h) paired with
images. Disfluencies and speed perturbation are added to the signal in order to sound more natural. Each speech signal (WAV) is paired with a JSON file containing exact timecode for each word/syllable/phoneme in the spoken caption. Such a corpus could be used for Language and Vision (LaVi) tasks including speech input or output instead of text.  Investigating multimodal learning schemes for unsupervised speech pattern discovery is also possible with this corpus, as demonstrated by a preliminary study conducted on a subset of the corpus (10h, 10k spoken captions).

\end{abstract}

%
%

\section{Introduction}

During the past few years, there has been an increasing interest in research gathering language and vision (LaVi) communities. This trend can be explained by the availability of multimodal corpora such as Flickr30k \cite{young-etal-2014}  or MSCOCO \cite{MSCOCO}, containing images and their captions in natural language text. LaVi systems also have benefited from the introduction of neural encoder-decoder approaches \cite{chorowski2015attention,KarpathyL15} that allow text to be generated from images or speech (or vice-versa), or to learn  joint embeddings of images and text \cite{WangLL15}.

After the pioneering work done at MIT 15 years ago \cite{Roy03}, these recent advances   have also fueled fundamental research in (grounded) language acquisition and understanding (linguistics, cognitive science, autonomous robotics) as well as in more applied research to solve tasks such as image captioning \cite{Bernardi:2016:ADG:3013558.3013571} or visual question answering \cite{WuTWSDH16}.

While many benchmark image datasets have been designed to assess the quality of machine-generated image descriptions (for a survey see \cite{Bernardi:2016:ADG:3013558.3013571}), speech modality  has been less focused on, probably because of the lack of large corpora for conducting such studies. In fact, except the augmentation of Flickr8k dataset (by collecting a corpus of 40,000 spoken captions using Amazon Mechanical Turk - AMT) proposed by \cite{Harwath2015} which lead to the following research papers \cite{HarwathG17,KamperSSL17},  very few researches on spoken language and vision were conducted so far. 

\vspace{0.1cm}

	\textbf{Paper contributions. }  

\vspace{0.1cm}

	This paper proposes a dataset that is an order of magnitude bigger than what is already available (600k spoken captions instead of 40k in  \cite{Harwath2015}). Our dataset  is built on top of MSCOCO \cite{MSCOCO} and contains more than 600,000  spoken captions (we will refer to it as SPEECH-COCO in the rest of this paper).
	We believe that this corpus may be useful for further Language and Vision (LaVi) tasks including speech input or output instead of text.  
	The data set is made available online\footnote{\url{https://zenodo.org/record/4282267}}.
	Moreover, speech data is perfectly annotated (does not rely on error-prone forced alignment) since the annotations are produced during the TTS process itself (at word, syllable and phone level).

\vspace{0.1cm}
	
	\textbf{Paper outline. }
	
	\vspace{0.1cm}

	This paper is organized as following. In section 2, we quickly review existing corpora usable for visually grounded language acquisition and motivates the choice of MSCOCO as a starting point. In section 3, we present our general methodology and describe the spoken caption generation itself while section 4 presents some analysis on SPEECH-COCO as well as its first use for an unsupervised word discovery task. Finally, section 5 concludes this work and gives some perspectives.

\section{Existing Corpora of Visually Grounded Speech}
\label{SOA}
	
	
	Flickr30K \cite{young-etal-2014} (an extension of Flickr8K \cite{Hodosh:2013:FID:2566972.2566993}) and MSCOCO \cite{MSCOCO} are now widely used in LaVi community. We can also mention the Visual Genome dataset, an ongoing effort to connect structured image concepts to language \cite{krishnavisualgenome}. For a broader survey on image captioning datasets, the reader may refer to \cite{Bernardi:2016:ADG:3013558.3013571}. 
	
	Flickr8K and Flickr30K contain images from Flickr with  approximately 8,000 and 30,000 images, respectively. The images in these two datasets were selected through user queries for specific objects and actions. Both contain five descriptions per image which were collected using a crowdsourcing platform (AMT).

The MSCOCO dataset currently consists of 123,287 images with five different descriptions per image. Images in this dataset are annotated for 80 object categories and bounding boxes around all instances in one of these categories are available for all images. The MSCOCO dataset has been widely used for automatic image captioning. Several extensions of MSCOCO are currently under development, including the addition of questions\&answers  for Visual Question Answering (VQA) task\footnote{\url{http://www.visualqa.org}} or FOIL-COCO (Find One mismatch between Image and Language caption) which consists of images associated with incorrect captions to challenge existing LaVi models\footnote{\url{https://foilunitn.github.io}}.
	
	Concerning spoken captions, AMT recordings were obtained from Flickr8K by \cite{Harwath2015} but only 40k captions are made available online\footnote{\url{https://groups.csail.mit.edu/sls/downloads/flickraudio/}}. In addition, time-coded annotations are obtained after alignment of speech data with transcripts through (error-prone) automatic forced-alignment. Very recently, spoken captions for MSCOCO were generated using Google TTS by \cite{Chrupala2017}. However, only one TTS voice was used limiting the speaker variability.

\section{SPEECH-COCO: Adding Spoken Captions to MSCOCO}
\label{TTS}

	\subsection{Our Starting Point:  MSCOCO}

We used Microsoft's Common Objects in Context (MSCOCO) \cite{MSCOCO} training and validation datasets as our starting point. In MSCOCO, each image is described by at least five descriptions written by humans. The images contain 91 common object categories (e.g. dog, elephant, bird, car, bicycle, air plane, etc.) from 11 super-categories (Animal, Vehicle, etc.), with 82 of them having more than 5K labelled instances. In total there are 123,287 images with captions (82,783 for training and 40,504 for validation). The test set is not available for download since it is used on an evaluation server for continuous benchmarking of image captioning systems. Consequently, 616,767 captions from 123,287 images are available for download.

	\subsection{Spoken Captions Generation}

Synthetic speech was generated for each caption using \emph{Voxygen}\footnote{\url{https://www.voxygen.fr}}, a commercial speech synthesis system, for 4 different UK voices (Paul, Elizabeth, Judith and Bronwen) and 4 different US voices (Phil, Bruce, Amanda and Jenny). It is important to note that this is corpus-based  concatenative speech synthesis \cite{Schwarz07a} and not parametric synthesis. So, for each speaker's voice, speech utterances are generated by concatenation of units mined from a large speech corpus (generally around 3000 sentences/speaker). This means that despite having little intra-speaker variability in our speech data, there is a realistic level of inter-speaker variability, as opposed to the small corpus proposed in \cite{Chrupala2017}.

On average, each caption comprises 10.79 tokens. The WAV files are on average 3.52 second long.

	\subsection{Adding Variability}

First, we applied speed perturbation on the spoken captions to introduce intra-speaker variability in the dataset. For this, we used the \textit{tempo} function of Sox\footnote{\url{http://sox.sourceforge.net/sox.html}} audio manipulation tool. This simple processing only changes speech rate while trying to keep the same pitch and spectral enveloppe. Our recordings were either accelerated (speed x 1.1), slowed down (speed x 0.9) or kept unchanged with equal probability of event.

Second, we added disfluencies to some of the captions so that they would sound more natural. We chose to only include fillers (such as "um", "uh", "er", "huh", "oh" and "ah"). It was shown that fillers often occur when speakers have to face an heavy processing load (when describing an image for example) and that they also act as coordinators \cite{bortfeld_disfluency}. Thus, adding fillers to some captions is legitimate and makes the corpus more realistic. The probability of adding a filler to any given caption was set to 0.3. The fillers were either added at the beginning, at the end or in the middle of a caption with equal probability of event. 

\section{Spoken Corpus Analysis}
\label{experiments}

	\subsection{Metadata and scripts}

We adopted the following naming convention for both the WAV and JSON files:
\begin{center}
\textit{imageID\_captionID\_Speaker\_DisfluencyPosition\_Speed}
\end{center}

\begin{figure*}[h!]
	\centering
\begin{center}
	\includegraphics[width=0.75\textwidth]{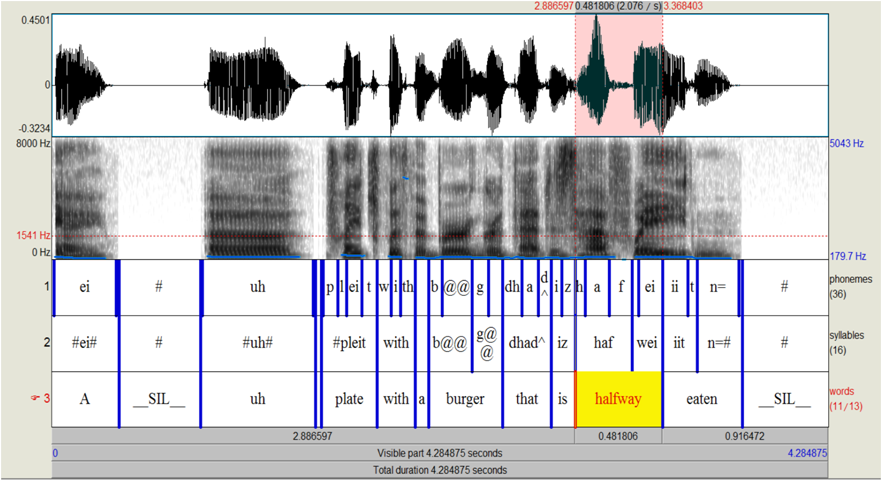}
\end{center}
	\caption{{\it Example of a spoken caption with different levels of annotation}}
	\label{praat}
\end{figure*}

Each WAV file is paired with a JSON file containing various information: timecode of each word in the caption, name of the speaker, name of the WAV file, etc. The JSON files have the following data structure:
\begin{lstlisting}[basicstyle=\fontsize{8}{8}\selectfont\ttfamily]	
{
	"duration": float, 
	"speaker": string, 
	"synthesisedCaption": string,
	"timecode": list, 
	"speed": float, 
	"wavFilename": string, 
	"captionID": int, 
	"imgID": int, 
	"disfluency": list
}
\end{lstlisting}

We also created a Python script\footnote{\url{https://github.com/William-N-Havard/SpeechCoco}} which handles the metadata, so that the user can easily make use of the corpus. The script has the following features: 

\begin{itemize}
	\item Aggregate all the information in the JSON files into a single SQLite database
	\item Find captions according to specific filters (name, gender and nationality of the speaker, disfluency position, speed, duration, and words in the caption). The script automatically builds the SQLite query. The user can also provide his own SQLite query.
	\item Find all the captions belonging to a specific image
	\item Parse the timecodes and have them structured
	\item Get the words, syllables and phonemes timecodes 
	\item Convert the timecodes to Praat TextGrid files (see Figure \ref{praat})
\end{itemize}

	\subsection{Analysis of intra and inter speaker variability}

To be sure that our corpus is realistic enough, we computed inter and intra speaker variability measures on a subset of 100 captions corresponding to 20 images and added 2 additional voices to our 8 synthetic voices for these captions: a real voice (William) and \textit{Google TTS} (gTTS).
We then extracted words that were repeated across sentences and analyzed their intra and inter speaker variability using Dynamic Time Warping (DTW) computed for all  occurrences of the same word by the same speaker (e.g. Amanda vs Amanda) and for all the occurrences of the same word pronounced by different speakers (e.g Amanda vs. Bronwen, Amanda vs. Bruce, etc.).

\begin{figure}[h!]
	\centering
\begin{center}
	\makebox[0.45\textwidth]{\includegraphics[width=0.5\textwidth]{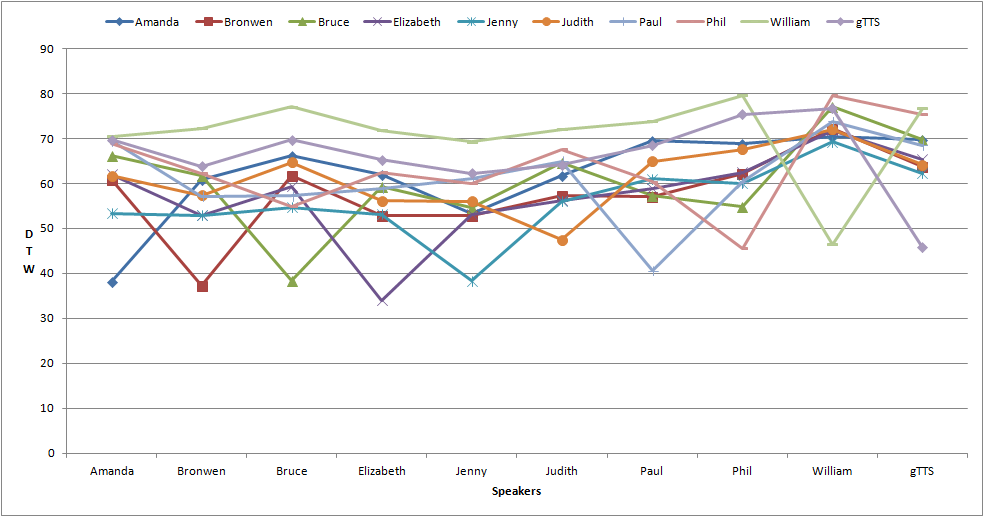}}
\end{center}
\caption{{\it Analysis of intra and inter speaker variability on the spoken captions (subset of 100 speech signals corresponding to 20 images)}}
	\label{DTW}
\end{figure}

As expected, intra speaker variability is lower than inter-speaker variability but we remark that there is not big differences between intra speaker variability of a real voice (William) and the intra speaker variability of a synthetic voice (Amanda for instance). Inter speaker variability is slightly greater between the real human voice (William) and the other synthetic voices, but inter speaker variability between  synthetic voices is still high, meaning our synthetic corpus should be difficult enough for tasks such as unsupervised term discovery (UTD) accross a collection of speech utterances from multiple speakers. This is what we intend to show in a preliminary experiment, in the next subsection.

	\subsection{Unsupervised Term Discovery (UTD)}

For UTD, we use the Zero Resource Toolkit (ZRTools \cite{jansen2011}). ZRTools uses segmental dynamic time warping (SDTW) to discover pairs of acoustically similar audio segments, and then uses graph clustering on overlapping pairs to form a hard clustering of the discovered segments.
Replacing each discovered segment with its unique cluster label, or pseudoterm, gives us a partial, noisy transcription, or pseudotext.

To evaluate UTD performance, we used the term discovery evaluation toolkit (TDE) \cite{tde}. UTD was done on a subset of the corpus (10,000 captions which represent approximately 10 hours of speech) and the results are reported in Table \ref{tab:UTD}. 

Our results confirm that the UTD task is still difficult even if our corpus is synthetic: we found very few \textit{gold tokens} (low precision and even lower recall\footnote{TDE only gives three significant figures after the decimal point. Since our results are very low, there were rounded up to 0.000}). There is very little difference between the global precision and the within-speaker precision for the matching task, suggesting that the clusters mainly consist of speech segments belonging to the same speaker. Within-speaker clusters also seem to be purer than the global clusters since the within-speaker precision for the grouping task is higher than the global precision. Grouping recall indicates that speech segments which should have been in the same clusters have been assigned to different clusters. Finally, the low coverage and matching recall clearly indicate that a lot of work has to be done in order to get the most out of the available data and that SDTW alone might not be enough to fully segment speech utterances. For instance, modeling prosodic features could be of great help. 

To summarize, this experiment with an off-the-shelf toolkit, shows that our synthetic corpus of spoken captions is difficult enough for UTD task.

Table \ref{tab:cluster} shows examples of clusters found by the UTD system along with the corresponding images of MSCOCO. As shown in Table \ref{tab:UTD}, segments rarely match single tokens. However, Table \ref{tab:cluster} shows that they often match frequent \textit{n-grams} ("a man riding", "fire hydrant" and "skiing down") and multiword expressions (MWE) such as "a bunch of bananas". As such, adding a new metric to TDE that would assess the quality of the segmentation according to \textit{gold chunks} and MWE would be interesting.

\section{Japanese translations}

A further augmentation of MSCOCO was created by \cite{miyazaki} where Japanese captions were collected using the same methodology as \cite{chen2015}. This corpus comprises 131,740 captions for 26,500 images. As the augmentation created by \cite{miyazaki} does not provided a Japanese version for all of the captions, we used Machine Translation (MT)\footnote{Using both Excite and Google's MT systems} to translate all the available English captions to Japanese. This allows SPEECH-COCO to be used for cross-modal studies using speech, images, text and translations.

\section{Conclusion}

In this paper, we have presented SPEECH-COCO, an extension of the MSCOCO image recognition and captioning dataset, which consists of more than 604 hours of speech. 

The addition of speech as a new modality enables SPEECH-COCO to be used in different fields of research including language acquisition, visually-grounded word discovery and keyword spotting, and semantic embedding using speech and vision.

This corpus has been used during the \textit{Jelinek Memorial Workshop on Speech and Language Technology} (JSALT) 2017 jointly organized by CMU and JHU (\textit{Speaking Rosetta Stone} team on  discovering grounded
linguistic units for languages without orthography).

We list below a list of topics that are as many LaVi tasks possible with SPEECH-COCO: 
\begin{itemize}
	\item spoken caption generation from images (generating speech from images), 
	\item visually-grounded spoken term discovery,
	\item cross-modal speech-image-text studies,
	\item image generation from speech input,
	\item simulating language documentation tasks where speech has been elicited from images,
	\item effect of visual context on computational language acquisition,
	\item (spoken) visual question answering,
	\item data augmentation by adding synthetic multi-modal data  to more naturalistic (but small) corpora.
\end{itemize}

\section{Acknowledgements}

We thank Voxygen for providing us their TTS server with UK and US English voices which allowed us to generate all our spoken captions.
We thank Hideto Kazawa from Google for providing automatic translations into Japanese. 

\begin{table*}
	\centering
	\resizebox{\textwidth}{!}{
	\begin{tabular}{|c|c|c|c|ccc|ccc|ccc|ccc|ccc|}
		\toprule
		\multicolumn{2}{|c|}{} & \multicolumn{5}{c|}{\textbf{MATCHING}} & \multicolumn{6}{c|}{\textbf{CLUSTERING}}      & \multicolumn{6}{c|}{\textbf{PARSING}} \\
		\midrule
		\multirow{2}[1]{*}{DTW thr.} & \multirow{2}[1]{*}{} & \multirow{2}[1]{*}{NED} & \multirow{2}[1]{*}{Coverage} & \multicolumn{3}{c|}{Matching} & \multicolumn{3}{c|}{Grouping} & \multicolumn{3}{c|}{Type} & \multicolumn{3}{c|}{Token} & \multicolumn{3}{c|}{Boundary} \\
		&       &       &       & P     & R     & F     & P     & R     & F     & P     & R     & F     & P     & R     & F     & P     & R     & F \\\hline
		\multirow{2}[0]{*}{0,86} & global & 24,7  & 8,9   & 33,2  & 0,1   & 0,2   & 16,5  & 48,9  & 24    & 0,5   & 0,1   & 0,2   & 0,3   & 0     & 0     & 28,3  & 3,2   & 5,8 \\
		& within & 25,1  & 8,8   & 33,8  & 0,6   & 1,2   & 20,3  & 49,4  & 28,4  & 0,4   & 0,1   & 0,1   & 0,3   & 0     & 0     & 27,9  & 3,2   & 5,6 \\\hline
		\multirow{2}[0]{*}{0,88} & global & 18,9  & 7,4   & 39,4  & 0,1   & 0,2   & 18,8  & 48,3  & 26,5  & 0,6   & 0,1   & 0,2   & 0,3   & 0     & 0     & 30,4  & 2,7   & 4,9 \\
		& within & 19,2  & 7,3   & 40,1  & 0,6   & 1,2   & 24,2  & 51,9  & 32,7  & 0,4   & 0,1   & 0,1   & 0,4   & 0     & 0     & 30,2  & 2,6   & 4,8 \\\hline
		\multirow{2}[0]{*}{0,9} & global & 17,2  & 5,9   & 44,9  & 0,1   & 0,2   & 22,8  & 64,2  & 32,1  & 0,6   & 0,1   & 0,2   & 0,4   & 0     & 0     & 32,4  & 2,1   & 4 \\
		& within & 15,5  & 5,8   & 46,7  & 0,5   & 1     & 28,4  & 51,7  & 35,8  & 0,6   & 0,1   & 0,1   & 0,4   & 0     & 0     & 32    & 2     & 3,8 \\\hline
	\end{tabular}%
	}
	\caption{Evaluation of the output of the UTD system on our synthetic corpus using TDE}
	\label{tab:UTD}%
\end{table*}%

\begin{table*}
	\centering
	\begin{tabular}{ | c | m{2.5cm} | m{5.5cm} | }
		\hline
		Pictures &  \begin{center}
			Transcription of segments in the cluster
		\end{center} & \begin{center}
			WAV files
		\end{center} \\\hline
		\begin{minipage}{8cm}
			\centering
			\includegraphics[height=0.2\textwidth]{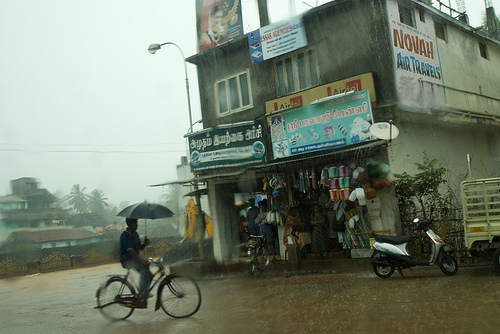}
			\includegraphics[height=0.2\textwidth]{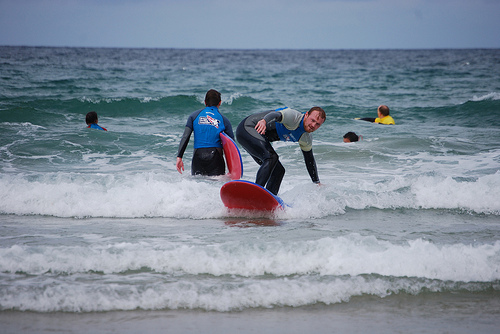}
			\includegraphics[height=0.2\textwidth]{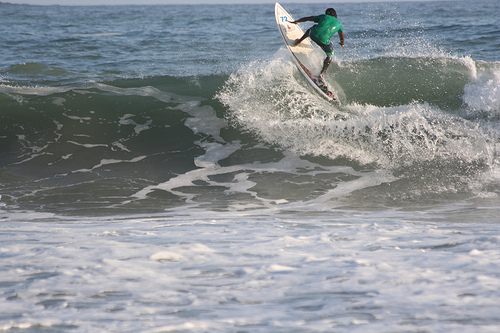}
			\includegraphics[height=0.2\textwidth]{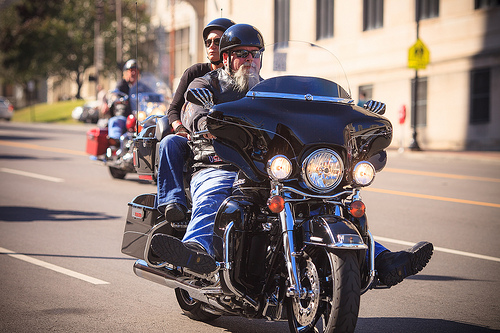}
			\includegraphics[height=0.2\textwidth]{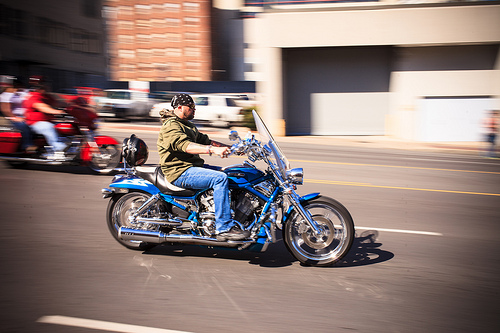}
			\includegraphics[height=0.2\textwidth]{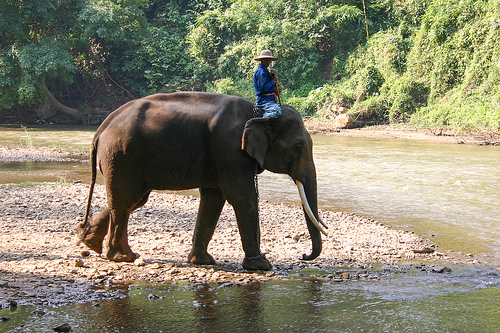}
		\end{minipage}
		&
		\begin{center}
			a man riding
		\end{center}
		& 
		\begin{center}
			1722\_448431\_Jenny\_End\_0-9
			131152\_736226\_Jenny\_None\_0-9
			1856\_609764\_Jenny\_None\_1-1
			132776\_377049\_Jenny\_None\_1-1
			264961\_171589\_Jenny\_None\_1-1
			131494\_321869\_Jenny\_Beginning\_0-9
			331366\_31230\_Jenny\_None\_1-0
			394941\_169848\_Jenny\_None\_0-9
			394840\_381007\_Jenny\_None\_1-1
		\end{center}
		\\ \hline
		\begin{minipage}{8cm}
			\centering
			\includegraphics[height=0.2\textwidth]{figures/247/247_1.jpg}
			\includegraphics[height=0.2\textwidth]{figures/247/247_2.jpg}
			\includegraphics[height=0.2\textwidth]{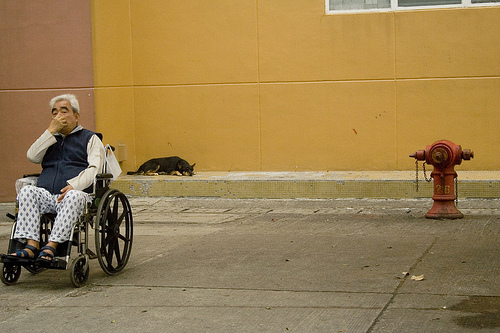}
			\includegraphics[height=0.2\textwidth]{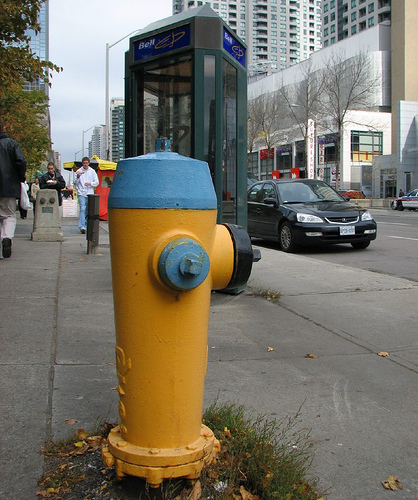}
			\includegraphics[height=0.2\textwidth]{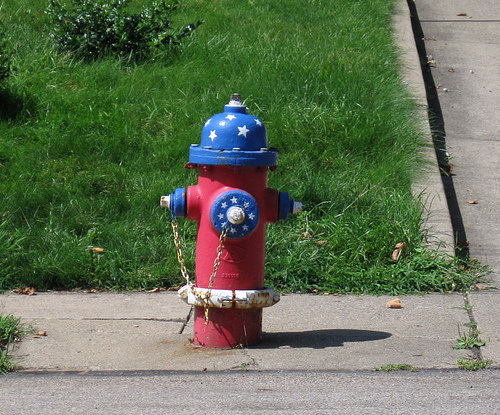}
			\includegraphics[height=0.2\textwidth]{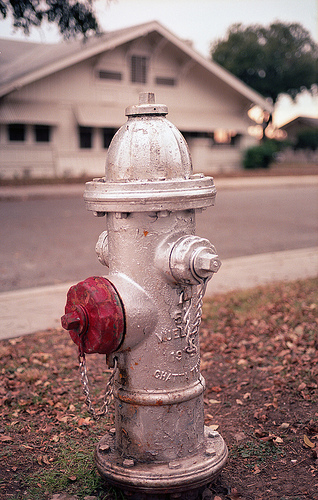}
		\end{minipage}
		&
		\begin{center}
			fire hydrant
		\end{center}
		& 
		\begin{center}
			357930\_221116\_Elizabeth\_None\_1-0
			262677\_222055\_Elizabeth\_End\_0-9
			469803\_313896\_Elizabeth\_End\_1-0
			425555\_174334\_Elizabeth\_End\_1-1
			264201\_230398\_Elizabeth\_None\_1-1
			131969\_332932\_Elizabeth\_None\_0-9
			262677\_251368\_Elizabeth\_None\_1-0
			264940\_680136\_Elizabeth\_None\_1-0
			197666\_684633\_Elizabeth\_None\_1-1
		\end{center}
		\\\hline
		\begin{minipage}{8cm}
			\centering
			\includegraphics[height=0.2\textwidth]{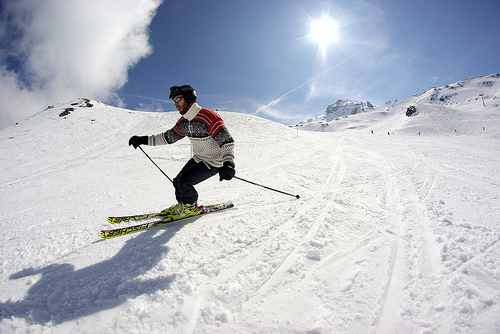}
			\includegraphics[height=0.2\textwidth]{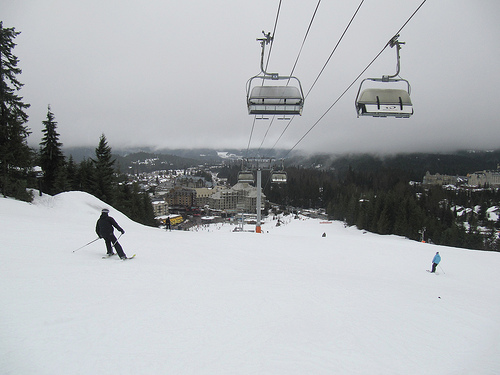}
			\includegraphics[height=0.2\textwidth]{figures/1266/1266_3.jpg}
		\end{minipage}
		&
		\begin{center}
			skiing down
		\end{center}
		& 
		\begin{center}
			415334\_686288\_Bronwen\_None\_1-1
			329827\_625065\_Bronwen\_None\_0-9
			134285\_298951\_Bronwen\_Beginning\_0-9
		\end{center}
		\\\hline
		\begin{minipage}{8cm}
		    \vspace{0.1cm}
			\centering
			\includegraphics[height=0.2\textwidth]{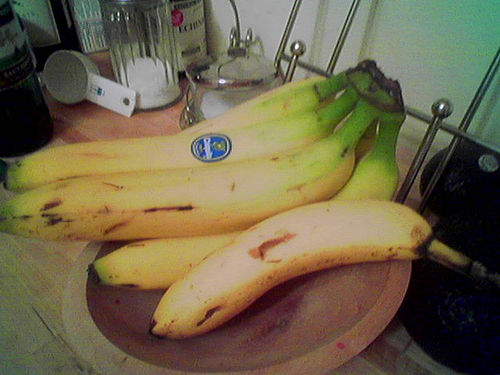}
			\includegraphics[height=0.2\textwidth]{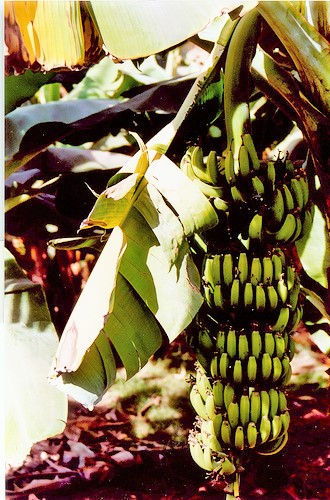}
			\vspace{0.1cm}
		\end{minipage}
		&
		\begin{center}
			a bunch of bananas
		\end{center}
		& 
		\begin{center}
			371683\_156894\_Judith\_None\_0-9
			1153\_459647\_Judith\_None\_0-9
		\end{center}
		\\\hline
	\end{tabular}
	\caption{Example of clusters found by the UTD from spoken captions only, along with the corresponding images of MSCOCO}
	\label{tab:cluster}
\end{table*}

\eightpt
\bibliographystyle{IEEEtran}
\bibliography{GLU2017}

\begin{thebibliography}{10}
\providecommand{\url}[1]{#1}
\csname url@samestyle\endcsname
\providecommand{\newblock}{\relax}
\providecommand{\bibinfo}[2]{#2}
\providecommand{\BIBentrySTDinterwordspacing}{\spaceskip=0pt\relax}
\providecommand{\BIBentryALTinterwordstretchfactor}{4}
\providecommand{\BIBentryALTinterwordspacing}{\spaceskip=\fontdimen2\font plus
\BIBentryALTinterwordstretchfactor\fontdimen3\font minus
  \fontdimen4\font\relax}
\providecommand{\BIBforeignlanguage}[2]{{%
\expandafter\ifx\csname l@#1\endcsname\relax
\typeout{** WARNING: IEEEtran.bst: No hyphenation pattern has been}%
\typeout{** loaded for the language `#1'. Using the pattern for}%
\typeout{** the default language instead.}%
\else
\language=\csname l@#1\endcsname
\fi
#2}}
\providecommand{\BIBdecl}{\relax}
\BIBdecl

\bibitem{young-etal-2014}
P.~Young, A.~Lai, M.~Hodosh, and J.~Hockenmaier, ``{From image descriptions to
  visual denotations: New similarity metrics for semantic inference over event
  descriptions},'' \emph{Transactions of the Association for Computational
  Linguistics}, vol.~2, pp. 67--78, 2014.

\bibitem{MSCOCO}
\BIBentryALTinterwordspacing
T.-Y. Lin, M.~Maire, S.~Belongie, J.~Hays, P.~Perona, D.~Ramanan,
  P.~Doll{\'a}r, and C.~L. Zitnick, ``Microsoft coco: Common objects in
  context,'' in \emph{European Conference on Computer Vision (ECCV)},
  Z{\"u}rich, 2014, oral. [Online]. Available:
  \url{/se3/wp-content/uploads/2014/09/coco_eccv.pdf, http://mscoco.org}
\BIBentrySTDinterwordspacing

\bibitem{chorowski2015attention}
J.~K. Chorowski, D.~Bahdanau, D.~Serdyuk, K.~Cho, and Y.~Bengio,
  ``{Attention-Based Models for Speech Recognition},'' in \emph{Advances in
  Neural Information Processing Systems (NIPS 2015)}, Montr{\'e}al, Canada,
  2015, pp. 577--585.

\bibitem{KarpathyL15}
\BIBentryALTinterwordspacing
A.~Karpathy and F.~Li, ``Deep visual-semantic alignments for generating image
  descriptions,'' in \emph{{IEEE} Conference on Computer Vision and Pattern
  Recognition, {CVPR} 2015, Boston, MA, USA, June 7-12, 2015}, 2015, pp.
  3128--3137. [Online]. Available:
  \url{https://doi.org/10.1109/CVPR.2015.7298932}
\BIBentrySTDinterwordspacing

\bibitem{WangLL15}
\BIBentryALTinterwordspacing
L.~Wang, Y.~Li, and S.~Lazebnik, ``Learning deep structure-preserving
  image-text embeddings,'' \emph{CoRR}, vol. abs/1511.06078, 2015. [Online].
  Available: \url{http://arxiv.org/abs/1511.06078}
\BIBentrySTDinterwordspacing

\bibitem{Roy03}
\BIBentryALTinterwordspacing
D.~Roy, ``Grounded spoken language acquisition: experiments in word learning,''
  \emph{{IEEE} Trans. Multimedia}, vol.~5, no.~2, pp. 197--209, 2003. [Online].
  Available: \url{https://doi.org/10.1109/TMM.2003.811618}
\BIBentrySTDinterwordspacing

\bibitem{Bernardi:2016:ADG:3013558.3013571}
\BIBentryALTinterwordspacing
R.~Bernardi, R.~Cakici, D.~Elliott, A.~Erdem, E.~Erdem, N.~Ikizler-Cinbis,
  F.~Keller, A.~Muscat, and B.~Plank, ``Automatic description generation from
  images: A survey of models, datasets, and evaluation measures,'' \emph{J.
  Artif. Int. Res.}, vol.~55, no.~1, pp. 409--442, Jan. 2016. [Online].
  Available: \url{http://dl.acm.org/citation.cfm?id=3013558.3013571}
\BIBentrySTDinterwordspacing

\bibitem{WuTWSDH16}
\BIBentryALTinterwordspacing
Q.~Wu, D.~Teney, P.~Wang, C.~Shen, A.~R. Dick, and A.~van~den Hengel, ``Visual
  question answering: {A} survey of methods and datasets,'' \emph{CoRR}, vol.
  abs/1607.05910, 2016. [Online]. Available:
  \url{http://arxiv.org/abs/1607.05910}
\BIBentrySTDinterwordspacing

\bibitem{Harwath2015}
D.~Harwath and J.~Glass, ``Deep multimodal semantic embeddings for speech and
  images,'' in \emph{IEEE Automatic Speech Recognition and Understanding
  Workshop}, Scottsdale, Arizona, USA, December 2015, pp. 237--244.

\bibitem{HarwathG17}
\BIBentryALTinterwordspacing
D.~F. Harwath and J.~R. Glass, ``Learning word-like units from joint
  audio-visual analysis,'' \emph{CoRR}, vol. abs/1701.07481, 2017. [Online].
  Available: \url{http://arxiv.org/abs/1701.07481}
\BIBentrySTDinterwordspacing

\bibitem{KamperSSL17}
\BIBentryALTinterwordspacing
H.~Kamper, S.~Settle, G.~Shakhnarovich, and K.~Livescu, ``Visually grounded
  learning of keyword prediction from untranscribed speech,'' \emph{CoRR}, vol.
  abs/1703.08136, 2017. [Online]. Available:
  \url{http://arxiv.org/abs/1703.08136}
\BIBentrySTDinterwordspacing

\bibitem{Hodosh:2013:FID:2566972.2566993}
\BIBentryALTinterwordspacing
M.~Hodosh, P.~Young, and J.~Hockenmaier, ``Framing image description as a
  ranking task: Data, models and evaluation metrics,'' \emph{J. Artif. Int.
  Res.}, vol.~47, no.~1, pp. 853--899, May 2013. [Online]. Available:
  \url{http://dl.acm.org/citation.cfm?id=2566972.2566993}
\BIBentrySTDinterwordspacing

\bibitem{krishnavisualgenome}
\BIBentryALTinterwordspacing
R.~Krishna, Y.~Zhu, O.~Groth, J.~Johnson, K.~Hata, J.~Kravitz, S.~Chen,
  Y.~Kalantidis, L.-J. Li, D.~A. Shamma, M.~Bernstein, and L.~Fei-Fei, ``Visual
  genome: Connecting language and vision using crowdsourced dense image
  annotations,'' 2016. [Online]. Available:
  \url{https://arxiv.org/abs/1602.07332}
\BIBentrySTDinterwordspacing

\bibitem{Chrupala2017}
G.~Chrupa{\l}a, L.~Gelderloos, and A.~Alishahi, ``Representations of language
  in a model of visually grounded speech signal,'' 2017, arXiv170201991 Cs.

\bibitem{Schwarz07a}
D.~Schwarz, ``Corpus-based concatenative synthesis : Assembling sounds by
  content-based selection of units from large sound databases,'' \emph{IEEE
  Signal Processing}, vol. 24-2, pp. 92--104, 2007.

\bibitem{bortfeld_disfluency}
H.~Bortfeld, S.~Leon, J.~Bloom, M.~Schober, and S.~Brennan,
  ``\BIBforeignlanguage{English}{Disfluency {Rates} in {Conversation}:
  {Effects} of {Age}, {Relationship}, {Topic}, {Role}, and {Gender}},''
  \emph{\BIBforeignlanguage{English}{Language and Speech}}, vol.~44, no.~2, pp.
  123--147, 2001.

\bibitem{jansen2011}
A.~Jansen and B.~Van~Durme, ``Efficient spoken term discovery using randomized
  algorithms,'' in \emph{Proceedings of ASRU}, 2011.

\bibitem{tde}
\BIBentryALTinterwordspacing
B.~Ludusan, M.~Versteegh, A.~Jansen, G.~Gravier, X.-N. Cao, M.~Johnson, and
  E.~Dupoux, ``\BIBforeignlanguage{en}{Bridging the gap between speech
  technology and natural language processing: an evaluation toolbox for term
  discovery systems},'' May 2014. [Online]. Available:
  \url{https://hal.inria.fr/hal-01026368/document}
\BIBentrySTDinterwordspacing

\bibitem{miyazaki}
\BIBentryALTinterwordspacing
T.~Miyazaki and N.~Shimizu, ``Cross-lingual image caption generation,'' in
  \emph{Proceedings of the 54th Annual Meeting of the Association for
  Computational Linguistics (Volume 1: Long Papers)}.\hskip 1em plus 0.5em
  minus 0.4em\relax Berlin, Germany: Association for Computational Linguistics,
  Aug. 2016, pp. 1780--1790. [Online]. Available:
  \url{https://www.aclweb.org/anthology/P16-1168}
\BIBentrySTDinterwordspacing

\bibitem{chen2015}
X.~Chen, H.~Fang, T.-Y. Lin, R.~Vedantam, S.~Gupta, P.~Doll{\'a}r, and C.~L.
  Zitnick, ``Microsoft coco captions: Data collection and evaluation server,''
  \emph{ArXiv}, vol. abs/1504.00325, 2015.

\end{thebibliography}
\end{document}